\documentclass{article}
\usepackage[numbers,sort&compress]{natbib}
\usepackage{spconf}
\usepackage{amsmath,amssymb,amsfonts}
\usepackage{algorithmic}
\usepackage{graphicx}
\usepackage{textcomp}
\usepackage{xcolor}
\usepackage{booktabs}
\usepackage{subfigure} 
\usepackage[misc]{ifsym}
\usepackage{multirow}
\usepackage{array}
\usepackage{colortbl}
\usepackage{nicematrix}
\usepackage{tikz}
\usetikzlibrary{calc, tikzmark}
\usepackage{makecell}

\usepackage[pagebackref=true,breaklinks=true,letterpaper=true,colorlinks, citecolor=blue,bookmarks=false]{hyperref}

\title{Label distribution learning via label correlation grid}
\twoauthors
{\normalsize Qimeng Guo, Liancheng Xu}
{\normalsize Shandong Normal University\\
 \normalsize	School of information science and Engineering}
{\normalsize Zhuoran Zheng, Xiuyi Jia}
{\normalsize Nanjing University of Science and Technology\\
 \normalsize	School of computer science and engineering}

\begin{document}
%
\maketitle
\begin{abstract}
Label distribution learning can characterize the polysemy of an instance through label distributions.
However, some noise and uncertainty may be introduced into the label space when processing label distribution data due to artificial or environmental factors. 
To alleviate this problem, we propose a \textbf{L}abel \textbf{C}orrelation \textbf{G}rid (LCG) to model the uncertainty of label relationships. 
Specifically, we compute a covariance matrix for the label space in the training set to represent the relationships between labels, then model the information distribution (Gaussian distribution function) for each element in the covariance matrix to obtain an LCG. 
Finally, our network learns the LCG to accurately estimate the label distribution for each instance. 
In addition, we propose a label distribution projection algorithm as a regularization term in the model training process.
Extensive experiments verify the effectiveness of our method on several real benchmarks.
\end{abstract}
\begin{keywords}
label distribution learning, covariance matrix, label correlation grid (LCG) 
\end{keywords}
\vspace{-2mm}
\section{Introduction}
\vspace{-2mm}
\label{sec:intro}
%
In contrast to the traditional single-label learning and multi-label learning, label distribution learning (LDL)~\cite{geng2016label} aims to characterize instance polysemy with labels of different description degrees. 
Specifically, LDL extends discrete labels into the form of a continuous probability distribution (description degree). 
The descriptive degree is deemed as the relevance of the label to the sample, and the sum of the descriptive degree of all labels is 1, showing the relationship between the instance and the sample. 
%
%
For example, a person's facial expression~\cite{zhao2021robust,DBLP:conf/icassp/SiWPX22,DBLP:journals/tii/ChenGXZYL22} may involve multiple emotions, including happiness, sadness, surprise, etc. 
LDL can convey such sophisticated and ambiguous semantic information.
%
%

%
Since researchers built label distribution datasets by using manual annotation and label enhancement algorithms~\cite{zhaofusion,DBLP:journals/corr/abs-2004-03104,DBLP:conf/icml/0009SLG20,DBLP:journals/tkde/XuLG21}, the label space of the dataset is usually disturbed by the introduction of uncertainty and noise.
On the one hand, the manually annotated datasets have a high degree of subjectivity and ambiguity. 
%
%
On the other hand, the inductive bias of the label enhancement algorithm cannot guarantee the accuracy of the label set.
The uncertainty of the label set further affects the modeling of label relations. 
To clear this obstacle, we model the uncertainty of label relations to boost the performance of the LDL algorithm.

Inspired by the idea of modeling the implicit label distribution proposed by~\cite{zhengLABEL}, our proposed model also implicitly characterizes the label relationship's uncertainty.
%
So far, we develop a novel approach to learning label distributions via a label correlation grid (LCG), which can model the relationships between labels and learn the implied uncertainty between pairs of labels. 
In particular, we first use an MLP (multi-layer perceptron) to extract the deep semantics of the input information in order to perform learning label distribution. 
Then, we use the predicted label distribution to compute a covariance matrix between the labels, with the help of which we generate an LCG.
Each cell in the grid is generated by sampling the element values in the covariance matrix using a Gaussian distribution function.
%
Note that this grid is constrained by the grid generated from the label space of the samples with the help of the $L_{1}$ norm.

In addition, since deep networks are prone to overfitting when learning tabular data, we propose a label distribution projection algorithm as a regularization term that performs on datasets with numerous labels.
The label distribution projection algorithm aims to extract the depth semantics of the label space (in this paper, we use KPCA~\cite{scholkopf1998nonlinear}), similar to the perceptual loss~\cite{gatys2016image}, which improves the generalization ability of the whole model.
Compared to traditional LDL algorithms, our model can execute acceleration on GPU (millisecond level speed on one inference), which is unparalleled competitive in terms of efficiency and accuracy.
Extensive experiments (including the standard accuracy experiment and the noise immunity experiment) have confirmed the effectiveness of our approach.
%
\textbf{Our main contribution} includes three folds:
\textbf{i)}: We propose a new method to learn label distributions via a label correlation grid that models the uncertainty of the label relationship.
\textbf{ii)}: We design a robust depth model that combines efficiency (ms level) and accuracy on a tabular dataset.
\textbf{iii)}: To alleviate the overfitting of the model, we design a label distribution projection algorithm on the dataset with a large number of labels.
\begin{figure*}[htbp]
	\centerline{\includegraphics[width=0.7\textwidth]{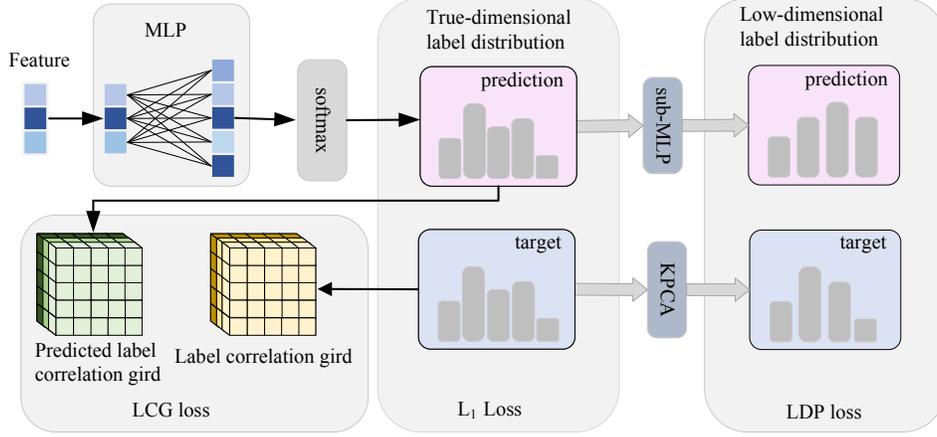}}
	\caption{\textbf{Our architecture}. Our algorithm aims to regress the label distribution of a sample using an MLP, where there are \textbf{three learning targets}, one is to learn the true label distribution, one is to learn the projection of the true label distribution, and one is to learn the LCG.}
	\label{fig}
	\vspace{-4mm}
\end{figure*} 

\vspace{-4mm}
\section{related work}
\label{sec:related work}
\vspace{-2mm}
Geng et al.~\cite{geng2016label} first proposed a new machine learning paradigm: LDL, which focuses on the relative importance of each label to an instance. 
More researchers have since devoted themselves to LDL \cite{DBLP:conf/emnlp/ZhouZZZG16,DBLP:conf/mm/ZhouXG15,DBLP:conf/ijcai/WangG19}.
In these researches, label correlation is usually applied to boost the generalization ability of the model.
LDL-LCLR~\cite{ren2019label} proposed to utilize both the global and local relevance among labels to train an LDL algorithm. 
Wang et al.~\cite{wang2021label} tried the underlying manifold structure of label distribution to encode the correlations among labels. 
Jia et al.~\cite{jia2021label} introduced a ranking loss in the LDL model to maintain the label ranking relationship. 
Qian et al.~\cite{qian2022feature} proposed an algorithm to generate the label correlations that uses neighborhood granularity to explore feature similarity while generating the label correlation with the correlation coefficient. 
In contrast to the explicit modeling learning algorithm described above, we focus on implicitly representing correlations between labels, while taking into account the uncertainty factor carried by the dataset itself.

\vspace{-2mm}
\section{PROPOSED METHOD}
\vspace{-2mm}
\label{sec:PROPOSED METHOD}
The architecture of our method is shown in Fig.~\ref{fig}. 
MLP learns the label distribution based on the deep semantics of the input features. 
Among them, we conduct three sub-learning targets, a key target is to learn \textit{the true label distribution} by using $L_{1}$ loss, followed by learning \textit{the LCG} and another one (label distribution projection loss) is to learn \textit{the deep semantics of the label distribution}.
%
Of these, the elements of the LCG come from the Gaussian distribution and the depth semantics of label distribution are from KPCA.
%
%

\noindent \textbf{Notations.}
Given a particular example, the goal of our method is to learn the degree to which each label describes that instance. 
Input matrix $X\in R^{m \times n}$, where $m$ is the number of instances and $n$ is the dimension of features. 
We define the \textit{i-th} instance in the dataset as $x_{i}$. 
The label distribution is defined $Y\in R^{m \times t} $, then the \textit{j-th }label is defined $y_{j} $. For each instance $x_{i}$, we define its label distribution $ D_{i}=\left\{ {d_{x_{i}}^{y_{1}},d_{x_{i}}^{y_{2}},... ,d_{x_{i}}^{y_{j}} }\right\} $ where $ d_{x_{i}}^{y_{j}} $ is the degree of description of the label $y_{j} $ for the sample $x_{i} $. 
The $ d_{x_{i}}^{y_{j}} $ is constrained by ${{d_{x_{i}}^{y_{j}}}}\in[0,1] $ and $\sum_{j=1}^{j}{d_{x_{i}}^{y_{j}}}=1$.

\noindent \textbf{Network Architecture.}
Early studies introduced neural networks into LDL, such as AA-BP~\cite{geng2016label}. 
We refer to this network architecture to regress a label distribution.
Firstly, the input instance $x_{i}$ is fed into an MLP with two linear layers. 
In the first hidden layer, we used 256 neurons to learn the latent features. 
%
The number of linear layer neurons in the last layer is the number of labels.
Note that the number of neurons is set to the power of 2 in favor of acceleration on the GPU shader.
Two non-linear layers (sigmoid and softmax) are added after the hidden layer and the output layer, respectively. 
The MLP has the ability to learn the latent semantic and global information of the input examples and generates accurate label distribution values. 
The predicted label description degree $\tilde{ d_{x_{i}}^{y_{j}} }$ of each instance $x_{i}$ is
\begin{equation}
	\begin{aligned}	
	\tilde{d_{x_{i}}^{y_{j}} }=\frac{\text{exp}(\text{MLP}(x_{i})) }{ \sum_{k=1}^{j}{\text{exp}(\text{MLP}(x_{k}))}}.	
		\label{equ1} 
	\end{aligned}
\end{equation}
We optimize the MLP by minimizing the $L_{1}$ distance between the predicted label distribution and the ground truth as follows:
\begin{equation}
	\begin{aligned}	
L_{\text{1}}\left(d_{i}, \tilde{d_{i}} \right)=\sum_{i=1}^{T}        \left| \left| {  d_{x_{i}}^{y_{j}}-\tilde{ d_{x_{i}}^{y_{j}}    }   } \right| \right|	,
		\label{equ7} 
	\end{aligned}
\end{equation}
where $\tilde{d_{i}}$ is the predicted label distribution and $d_{i}$ is the true label distribution, and \textit{T} denotes the number of labels.
The following describes some key regularization terms.

\noindent  \textbf{Label Correlation Grid.}
Implicit representation~\cite{zhengLABEL} has nice properties due to its powerful expressiveness and for this reason, we develop the scheme of implicitly representing label relations.
To robustly model label uncertainty, we build a tensor that can accommodate high-dimensional information. 
%
Summing up the above thesis, we propose an LCG $y\_grid \in R^{t \times t \times t}$. 
$y\_grid$ is initially composed of a covariance matrix computed from the label space, where each element is sampled by a Gaussian function to create a vector.
The Gaussian distribution has two hyperparameters and we use each element of the covariance matrix as the mean and the variance is set fixed at 0.5.
We also try other values \{0.1, 0.2, ... ,0.9\} as the variance, but the results are not good as seen from the extensive experimental results.
%
%
%
The LCG is shown as:
\begin{equation*}
	\begin{bNiceMatrix}[name=a]
		y_{1}^{1} & \cellcolor{blue!20} y_{1}^{2} &\cdots&y_{1}^{t} \\
		y_{2}^{1} & y_{2}^{2} &\cdots& y_{2}^{t} \\
		\vdots&\vdots& \ddots&\vdots \\
		y_{t}^{1} & y_{t}^{2}&\cdots & y_{t}^{t}
	\end{bNiceMatrix}\qquad
	\begin{bNiceMatrix}[name=b]
	q_{1}, q_{2}, ..., \tikzmarknode[anchor=south]{upper right}{\strut q_{l}}
	\end{bNiceMatrix}
	\tikz[remember picture, overlay]
	\draw[->, red] (a-1-2) to[in=180, out=10] (b-1-1);
\end{equation*}
in which $y_{i}^{j}=[q_{1},q_{2},...,q_{l}]$ indicates the information between the \textit{i-th} label and the \textit{j-th} label.
The final LCG is optimized by calculating the $L_{1}$ distance.
A large number of papers use Gaussian priors for uncertainty estimation of unknown information. So we use the Gaussian process regression~\cite{wilson2011gaussian} to generate the description information.
%

\noindent \textbf{Label Distribution Projection Algorithm.} 
To boost the performance of the model, we propose a label distribution projection algorithm as a regularization term in the network training process.
We use a sub-MLP to aggregate feature information into a lower dimension in the network. 
The sub-MLP consists of one linear layer with sigmoid and softmax as activation functions. 
For example,  the 68-dimensional raw label distribution is mapped to a the 32-dimensional label distribution space by sub-MLP for the \textit{Human Gene} dataset.
The formulaic expression form is as follows:
\begin{equation}
	\begin{aligned}	
 \tilde{ d_{x_{i}}^{y_{j}}} '= \text{softmax}(\text{sub-MLP}(d_{x_{i}}^{y_{j}})).	
		\label{equ5} 
	\end{aligned}
\end{equation}

For true low-dimensional label distributions, we use Kernel-based principal component analysis (KPCA) to project the raw true label distribution. 
KPCA draws on a nonlinear approach to extract the abstract semantics of the label distribution, and in addition, has the effect of reducing noise.
The lower label distribution is calculated as ${d_{x_{i}}^{y_{j}}}'$.
We use $L_{1}$ to minimize the output of the network and the true low-dimensional label distribution.

\noindent \textbf{Loss Function.} 
%
The network parameters are optimized by combining $L_{1}$, projection loss ($L_{ldp}$) and label correlation grid loss ($L_{lcg}$).
Combined with the content of the above, our target function is constructed as follows:
\begin{equation}
	\begin{aligned}	
Loss=\lambda_{1} \times L_{1}+\lambda_{2} \times L_{ldp}+\lambda_{3} \times L_{lcg},
		\label{equ4} 
	\end{aligned}
\end{equation}
where $\lambda_{1}$, $\lambda_{2}$ and $\lambda_{3} $ are parameters. In the four label distribution datasets, we select  $\lambda_{1}$, $\lambda_{2}$ and $\lambda_{3} $ to be 1, 0.1 and 0.05, respectively.
\begin{table}[htbp] \footnotesize 
	\centering
	\caption{Statistics of four real-world datasets.}
    \setlength{\tabcolsep}{5mm}{
	\begin{tabular}{c|ccc}
		\toprule
		Dataset & Examples & Features & Labels \\
		\midrule
		Human Gene & 17892 & 36    & 68 \\
		Natural Scene & 2000  & 294   & 9 \\
		Yeast-alpha  & 2465   & 24   & 18 \\
		Movie & 7755  & 1869  & 5 \\
		\bottomrule
	\end{tabular}%
    }
	\label{tab:1}%
\end{table}%
\begin{table}[htbp] \footnotesize 
	\centering
	\caption{Parameters on the four datasets.}
	\begin{tabular}{c|cccc}
		\toprule
		Dataset & Batch size & Learning rate  & Epoch & LDP loss \\
		\midrule
		Human Gene & 500   & 0.0005 & 400   & 32 \\
		Natural Scene & 500   & 0.0005 & 300   & \textbf{-} \\
		Yeast-alpha  & 500   & 0.0005 & 400   & 16 \\
		Movie & 2000  & 0.0005 & 100   & \textbf{-} \\
		\bottomrule
	\end{tabular}%
	\label{tab:2}%
\end{table}%
\vspace{-2mm} 
\section{EXPERIMENTS}
\label{sec:EXPERIMENTS}
\textbf{Datasets.}
To validate the efficiency of our model, we conduct experiments on four LDL datasets. 
Details of the datasets from Wang et al.~\cite{wang2021label} are presented in Table~\ref{tab:1}. 
The datasets involve different types, including biological experiment datasets, natural scene datasets, movie category datasets, and large-scale biomedical research datasets.  
The evaluation is based on six indicators proposed by~\cite{geng2016label}, including Chebyshev distance (Chebyshev), Clark distance (Clark), Canberra distance (Canberra), KL divergence (K-L), Cosine similarity (Cosine), and Intersection similarity (Intersection).

\noindent \textbf{Comparative Algorithms.}
We compared the results of our model with five LDL algorithms: DDH-LDL~\cite{DBLP:journals/spl/ZhangZLX22}, BFGSLLD \cite{geng2016label}, LDL-LRR~\cite{jia2021label}, LDLSF~\cite{DBLP:conf/ijcai/RenJLCL19} and LALOT~\cite{DBLP:conf/aaai/ZhaoZ18a}.
Except for DDH-LDL, the indicators for all datasets of the models are from ~\cite{jia2021label}. Our experiments in DDH-LDL are based on the tips in the code website corresponding to the article.\\
\begin{table*}[htbp]
	\centering
	\caption{Experimental results on four datasets and the best results are bolded.}
	\resizebox{\textwidth}{50mm}{
	\begin{tabular}{c|c|cccccc}
	\toprule
	\textbf{Dataset} & \textbf{Algorithm} & \multicolumn{1}{c|}{\textbf{Chebyshev $\downarrow$}} & \multicolumn{1}{c|}{\textbf{Clark $\downarrow$}} & \multicolumn{1}{c|}{\textbf{Canberra $\downarrow$}} & \multicolumn{1}{c|}{\textbf{K-L $\downarrow$}} & \multicolumn{1}{c|}{\textbf{Cosine $\uparrow$}} & \textbf{Intersection $\uparrow$} \\
	\midrule
    \multirow{6}[2]{*}{Movie} & Ours  & \textbf{0.1145$\pm$0.0073} & \textbf{0.5054$\pm$0.0027} & \textbf{0.9746$\pm$0.0055} & 0.1178$\pm$0.0029 & \textbf{0.9446$\pm$0.0092} & \textbf{0.8384$\pm$0.0090} \\
& DDH-LDL & 0.1543$\pm$0.0634 & 0.6427$\pm$0.2526 & 1.2094$\pm$0.0051 & 0.1451$\pm$0.0127 & 0.9026$\pm$0.0048 & 0.7915$\pm$0.0913 \\
& LDL-LRR & 0.1151$\pm$0.0016 & 0.5235$\pm$0.0065 & 0.9996$\pm$0.0093 & \textbf{0.0982$\pm$.0031} & 0.9353$\pm$0.0020 & 0.8355$\pm$0.0023 \\
& BFGS-LLD & 0.1397$\pm$0.0223 & 0.5856$\pm$0.0494 & 1.1309$\pm$0.1098 & 0.1378$\pm$0.0373 & 0.9108$\pm$.0227 & 0.8053$\pm$0.0263 \\
& LALOT & 0.2854$\pm$0.0903 & 1.431$\pm$0.9006 & 2.9023$\pm$0.8996 & 3.7216$\pm$0.9542 & 0.7021$\pm$0.1415 & 0.6015$\pm$0.2114 \\
& LDLSF & 0.1271$\pm$0.0027 & 0.6227$\pm$0.0112 & 1.0166$\pm$0.0158 & 0.1954$\pm$0.0257 & 0.9220$\pm$0.0028 & 0.8180$\pm$0.0034 \\
\midrule
\multirow{6}[2]{*}{Yeast-alpha} & Ours  & \textbf{0.0133$\pm$0.0025} & 0.2100$\pm$0.0083 & \textbf{0.6708$\pm$0.2624} & 0.0062$\pm$0.0194 & \textbf{0.9947$\pm$0.0099} & \textbf{0.9626$\pm$0.0056} \\
& DDH-LDL & 0.1627$\pm$0.0529 & 0.5228$\pm$0.1937 & 1.0236$\pm$0.3634 & 0.1177$\pm$0.0638 & 0.8794$\pm$0.0936 & 0.8128$\pm$0.9355 \\
& LDL-LRR & 0.0134$\pm$0.0002 & \textbf{0.2093$\pm$0.0031} & 0.6791$\pm$0.0100 & \textbf{0.0054$\pm$0.0001} & 0.9946$\pm$0.0001 & 0.9625$\pm$0.0005 \\
& BFGS-LLD & 0.0135$\pm$0.0001 & 0.2110$\pm$0.0030 & 0.6865$\pm$0.0106 & 0.0055$\pm$0.0001 & 0.9949$\pm$0.0001 & 0.9629$\pm$0.0006 \\
& LALOT & 0.0165$\pm$0.0003 & 0.2608$\pm$0.0043 & 0.8544$\pm$0.0150 & 0.0084$\pm$0.0003 & 0.9917$\pm$0.0003 & 0.9526$\pm$0.0008 \\
& LDLSF & 0.0139$\pm$.0002 & 0.2164$\pm$0.0021 & 0.6874$\pm$0.0111 & 0.0058$\pm$0.0001 & 0.9943$\pm$0.0001 & 0.9613$\pm$.00004 \\
\midrule
\multirow{6}[2]{*}{Natural Scene} & Ours  & 0.3298$\pm$0.0020 & \textbf{1.8424$\pm$0.0004} & \textbf{4.3583$\pm$0.0125} & 2.332$\pm$0.00178 & \textbf{0.7607$\pm$0.0075} & \textbf{0.5749$\pm$0.0078} \\
& DDH-LDL & 0.4051$\pm$0.0022 & 2.1792$\pm$0.0037 & 5.5492$\pm$1.3133 & 1.6338$\pm$0.4791 & 0.5603$\pm$0.0067 & 0.4228$\pm$0.0055 \\
& LDL-LRR & \textbf{0.2997$\pm$0.0070} & 2.4268$\pm$0.0106 & 5.7143$\pm$0.0126 & \textbf{0.7314$\pm$0.0070} & 0.7532$\pm$0.0044 & 0.5670$\pm$0.0022 \\
& BFGS-LLD & 0.3511$\pm$0.0158 & 2.4756$\pm$0.0217 & 5.8445$\pm$0.0932 & 0.9067$\pm$0.0556 & 0.6765$\pm$0.0211 & 0.4778$\pm$0.0196 \\
& LALOT & 0.3873$\pm$0.0033 & 2.4999$\pm$0.0030 & 5.9962$\pm$0.0251 & 1.2529$\pm$0.0145 & 0.5473$\pm$0.0046 & 0.3600$\pm$0.0029 \\
& LDLSF & 0.3196$\pm$0.0092 & 2.4622$\pm$0.0159 & 6.6739$\pm$0.0139 & 0.9981$\pm$0.0328 & 0.7297$\pm$0.0074 & 0.5397$\pm$0.0074 \\
\midrule
\multirow{6}[2]{*}{Human gene} & Ours  & \textbf{0.0528$\pm$0.0074} & \textbf{2.1000$\pm$1.245} & 14.3673$\pm$0.1252 & \textbf{0.2271$\pm$0.0347} & \textbf{0.8352$\pm$0.0082} & \textbf{0.7849$\pm$0.0085} \\
& DDH-LDL & 0.0597$\pm$0.0752 & 2.8830$\pm$0.8514 & 19.9391$\pm$0.9627 & 0.3815$\pm$0.3870 & 0.7313$\pm$0.8355 & 0.6878$\pm$0.8845 \\
& LDL-LRR & 0.0532$\pm$0.0011 & 2.1114$\pm$0.0122 & \textbf{13.5681$\pm$0.0025} & 0.2365$\pm$0.0049 & 0.8346$\pm$0.0020 & 0.7844$\pm$0.0014 \\
& BFGS-LLD & 0.0539$\pm$0.0009 & 2.1270$\pm$0.0141 & 14.5633$\pm$0.1107 & 0.2398$\pm$0.0038 & 0.8328$\pm$0.0018 & 0.7828$\pm$0.0014 \\
& LALOT & 0.0573$\pm$0.0012 & 4.0703$\pm$3.6917 & 17.8198$\pm$3.7167 & 0.2956$\pm$0.0059 & 0.8055$\pm$0.0024 & 0.7578$\pm$0.0016 \\
& LDLSF & 0.0533$\pm$0.0009 & 2.1295$\pm$0.0209 & 14.5681$\pm$0.0055 & 0.2395$\pm$0.0054 & 0.8332$\pm$0.0028 & 0.7828$\pm$0.0022 \\

		\bottomrule
	\end{tabular}%
}
	\label{tab:3}%
\end{table*}%
\noindent \textbf{Experimental Setting.}
For our algorithm, we put the customized selection of parameters for each dataset in Table~\ref{tab:2}. 
The parameters that may be used include batch size, learning rate, and epoch. 
Since the label space of the three datasets \textit{Natural Scene}, \textit{Movie} is small, there is not enforced LDP loss function in the training phase.
%
Here, 32 indicates that we squeeze the raw label dimension from 68 to 32 using KPCA, followed immediately by normalization using softmax.
Besides, we use the PyTorch framework and AdamW optimizer to train and test a deep network on the GPU shader.

\noindent \textbf{Results and Discussion.}
The experimental results for each dataset are summarized in Table~\ref{tab:3}, where we take the results of the 10 times 5-fold cross-validation. 
The experimental results are reported in the form of ``mean$\pm$std''. 
$\uparrow$ indicates that the higher the value of the metric, the better.
$\downarrow$ indicates that the smaller the value of the metric, the better.

Our algorithm performs optimally on six parameters across the four datasets, particularly on the metric \textit{Clark}. 
Our approach also achieves competitive results in the remaining metrics. 
This is mainly due to the ability of the algorithm to expand label relationships and mine feature semantic information.  
Besides, we observe that algorithms that exploit label relations, like LDL-LRR and ours, generally outperform other algorithms, which demonstrates the importance of label relations in label distribution. 
%
Our method has a better performance compared to LDL-LRR by taking into account the uncertainty of the label relationship.
Although our modeling has an expensive cost, a GPU acceleration can be enforced in the PyTorch to alleviate this problem.
We evaluate the inference speed of our model on CPU (AMD 8-core processor) and GPU (RTX3080) as \textbf{22ms} and \textbf{1ms}, respectively.

\noindent \textbf{Ablation Studies.}
We conduct ablation studies to demonstrate the effectiveness of LCG and label distribution projection algorithms on dataset \textit{Human gene}.  
The results of simple ablation experiments are shown in Table~\ref{tab:4}. 
w/o grid is an algorithm for removing the LCG and w/o $ L_{dpa}$ is defined when the loss term $ L_{dpa}$  is removed. 
The variation of the data illustrates the effectiveness of the LCG and the label distribution projection algorithms. 
We conduct 10 times 5-fold cross-validation on the dataset of the ablation studies.
The \textcolor{gray}{gray} markers represent optimal performance.

\noindent \textbf{Noise Interference.}
To verify the robustness of our algorithm against the noise and uncertainty in label space, we experimented on the dataset \textit{Movie} with Gaussian noise.
We synthesize a dataset on \textit{Movie} by Gaussian function to enforce different degrees of noise, and the labels with noise are finally normalized by softmax.
%
For label space on the dataset, the variance of the Gaussian distribution function is selected from (0.1, 0.2, ..., 1). 

\begin{table}[htbp]
	\centering
	\caption{Results on Ablation studies.}
	\resizebox{1\columnwidth}{!}{
		\begin{tabular}{c|ccc}
			\toprule
			\textbf{Measures
			} & \multicolumn{1}{c|}{Ours} & \multicolumn{1}{c|}{w/o grid } & w/o  $ L_{dpa}$ \\
			\midrule
			Chebyshev $\downarrow$ & \cellcolor{lightgray}0.0528$\pm$0.0074 & 0.5346$\pm$0.0021 & 0.5349$\pm$0.0132 \\
			Clark $\downarrow$ & \cellcolor{lightgray}2.1000$\pm$1.2455 & 2.1011$\pm$1.1352 & 2.1073$\pm$0.0047 \\
			Canberra $\downarrow$ & \cellcolor{lightgray}14.3673$\pm$0.1252 & 14.3798$\pm$0.1368 & 14.3689$\pm$0.0054 \\
			K-L $\downarrow$ & \cellcolor{lightgray}0.2271$\pm$0.0347 & 0.2274$\pm$0.0052 & 0.2283$\pm$0.0032 \\
			Cosine $\uparrow$ & \cellcolor{lightgray}0.8352$\pm$0.0082 & 0.8350$\pm$0.0086 & 0.8313$\pm$0.0861 \\
			Intersection $\uparrow$ & \cellcolor{lightgray}0.7849$\pm$0.0085 & 0.7845$\pm$0.0074 & 0.7841$\pm$0.0015 \\
			\bottomrule
		\end{tabular}%
	}
\vspace{-5mm}
	\label{tab:4}%
\end{table}%
It is widely believed that the higher the variance, the more dispersed the data, and the more noise. 
%
%
We conduct 10 times 5-fold cross-validation on the synthetic dataset. 
Our algorithm yield 0.1257$\pm$0.0071 (Chebyshev), 0.5713$\pm$0.0027 (Clark), 1.0854$\pm$0.0055 (Canberra), 0.1452$\pm$0.0033 (K-L), 0.9250$\pm$0.0092 (Cosine), 0.8203$\pm$0.0037 (Intersection) respectively on the data set (the variance is 1) with Gaussian noise has a variance of 1 and mean of 0. 
However, the accuracy of the label distribution predicted by other algorithms decreases by \textbf{50\% on average}, and our model is still very competitive.

\vspace{-2mm}
\section{CONCLUSION}
\vspace{-2mm}
Ambiguity and uncertainty of label space in LDL become major challenges, limiting the performance of LDL models.
In this paper, we propose a new algorithm for label distribution by using an LCG. 
In standard evaluation experiments, our method achieves competitive results, especially on the metric of vector similarity rather than distance.
Furthermore, our method has a striking performance on the mini-test of noise resistance, thanks to the gains from uncertainty modeling.
Our model can execute GPU acceleration on processing LDL data, balancing accuracy and efficiency.
%
%
%


\vfill\pagebreak


\bibliographystyle{IEEEbib}
\bibliography{refs}

\end{document}